# Introduction to the Soar Cognitive Architecture[1]

John E. Laird[2]

May 8, 2022

This paper is the recommended initial reading for a functional overview of Soar, version 9.6. It supplements more in-depth descriptions of Soar, such as the Soar book (Laird, 2012); the Soar Tutorial (Laird, 2017), which provides a step-by-step introduction to programming Soar; and the Soar Manual (Laird et al., 2017), which gives complete details of Soar's operation and use.

Soar is meant to be a general cognitive architecture (Langley et al., 2009) that provides the fixed computational building blocks for creating AI agents whose cognitive characteristics and capabilities approach those found in humans (Laird, 2012; Newell, 1990). A cognitive architecture is not a single algorithm or method for solving a specific problem; rather, it is the task-independent infrastructure that learns, encodes, and applies an agent's knowledge to produce behavior, making a cognitive architecture a software implementation of a general theory of intelligence. One of the most difficult challenges in cognitive architecture design is to create sufficient structure to support coherent and purposeful behavior, while at the same time providing sufficient flexibility so that an agent can adapt (via learning) to the specifics of its tasks and environment. The structure of Soar is inspired by the human mind and as Allen Newell (Newell, 1990) suggested over 30 years ago, it attempts to embody a unified theory of cognition.

Soar was originally developed in the early 1980s as an architecture to support multi-task, multi-method problem solving. Its evolution has led to new modules and capabilities, including reinforcement learning, episodic and semantic memory, and the spatial visual system (Laird & Rosenbloom, 1996; Laird, 2012). It is a freely available open-source project (https://soar.eecs.umich.edu/). Over the years, a wide range of agents has been developed in Soar, many of which are briefly described in the appendix. These include agents embodied in real-world robots, computer games, and large-scale distributed simulation environments. These agents incorporate combinations of real-time decision-making, planning, natural language understanding, metacognition, theory of mind, mental imagery, and multiple forms of learning. Soar's focus has mainly been on AI agents, but it has been used for detailed modeling of human behavior (Stearns, 2021; Schatz et al., 2022).

A key hypothesis of Soar is that there are sufficient regularities above the neural level to capture the functionality of the human mind. Thus, the majority of knowledge representations in Soar are symbol structures, with architecturally maintained numeric metadata biasing the retrieval and learning of those structures. Soar also includes supports non-symbolic reasoning through the spatial visual system, which is an interface between perception and working memory. There is no commitment to a specific underlying implementation level, such as using simulated neurons. Our commitment is to an efficient and portable implementation, which happens to be in C/ C++ (a Java version duplicates the functionality of Soar 9.3). All agents developed in Soar use the same architecture and users never program in C/C++.

---

[1] There are currently no plans to formally publish this paper. I will continually update this paper and make it available through the Soar website, arXiv, and similar venues.

[2] Co-director of the Center for Integrated Cognition (john.laird@cic.iqmri.org); Professor of Computer Science and Engineering, University of Michigan (laird@umich.edu).

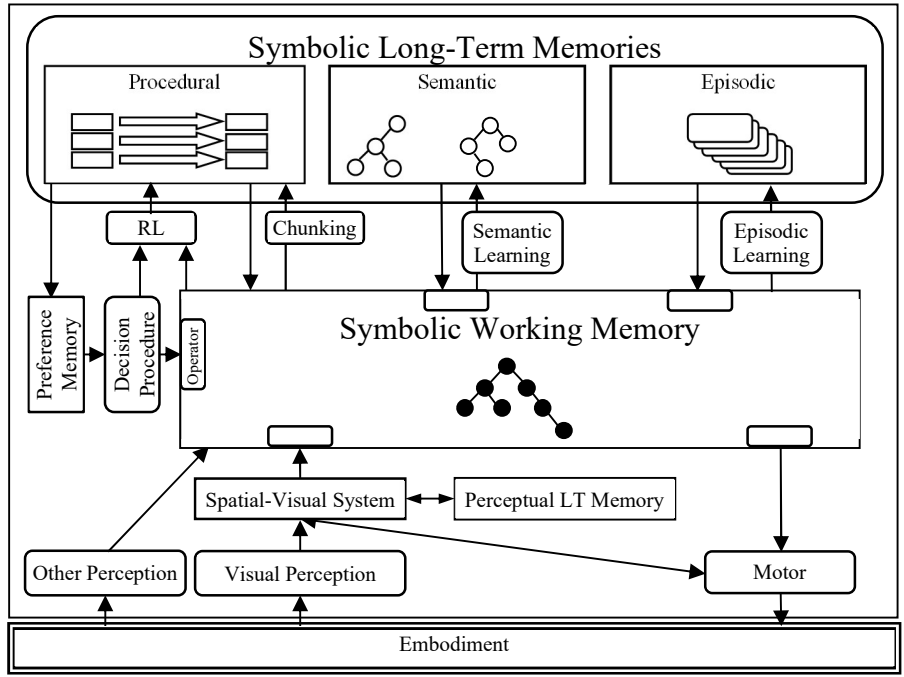

Figure 1: Structure of Soar memories, processing modules, learning modules and their connections.

Soar shares many characteristics with other cognitive architectures (Kotseruba & Tsotsos, 2020). The similarities shared by Soar, ACT-R (Anderson et al., 2004), and Sigma (Rosenbloom et al., 2016) led to the development of the Common Model of Cognition (Laird et al., 2017), which attempts to provide an abstract specification for cognitive architectures developed for human-like cognition. Soar and ACT-R are the two oldest and most widely used cognitive architectures, with ACT-R emphasizing cognitive modeling and connections to brain structures (Anderson, 2006). Laird (2021) provides a detailed analysis and comparison of Soar and ACT-R.

Section 1 provides an abstract overview of the architectural structure of Soar as depicted in Figure 1, including its processing, memories, learning modules, their interfaces, and the representations of knowledge used by those modules. Sections 2-8 describe the processing supported by those modules, including decision making (Section 2), impasses and substates (Section 3), procedure learning via chunking (Section 4), reinforcement learning (Section 5), semantic memory (Section 6), episodic memory (Section 7), and spatial-visual reasoning (Section 8). Sections 9-10 provide a review of the levels of decision making and variety of learning in Soar (Section 9), and analysis of Soar as an architecture supporting general human-level AI (Section 10). Following the references is an appendix that contains short descriptions of recent Soar agents and a glossary of the terminology we use in describing Soar.

## 1. Structure of Soar

Figure 1 shows the structure of Soar, which consists of interacting task-independent modules. There are short-term and long-term memories, processing modules, learning mechanisms, and interfaces between them. Working memory maintains an agent's situational awareness, including perceptual input, intermediate reasoning results, active goals, hypothetical states, and buffers for interacting with semantic memory, episodic memory, the spatial-visual system (SVS), and the motor system.

Long-term symbolic knowledge is stored in three memories (top of the figure). These include procedural memory (skills and "how-to" knowledge); semantic memory (facts about the world and the agent); and episodic memory (memories of experiences). Procedural knowledge drives behavior by responding to the contents of working memory and making modifications to it. Procedural memory implements purely internal reasoning, as well as initiating retrievals from semantic and episodic memory of knowledge into working memory. Procedural knowledge can also initiate actions with either SVS for spatial/imagery-based reasoning or with the motor system for action in the world. Automatic learning mechanisms are associated with procedural and episodic memories.

A simple example often helps people understand the basic operation of Soar, at least at an abstract level. I will ask you to perform a simple task, and in parallel, I will explain how it is performed in Soar by an agent implemented by James Boggs. This description is necessarily abstract and will not completely explain the details of processing, but it will give you an overview of how information flows through Soar and how procedural memory drives reasoning. To start with, imagine the word "WOW." This works better if you hear this verbally, but you should get the overall idea.

In this case, the instructions are verbal and are transformed by "other perception" into a symbolic representation of words. Procedural knowledge for language understanding tests for the existence of words and sentences, parses the instructions, retrieves the meaning of each word from semantic memory, and creates an internal representation of the meaning of the sentence. In this case, a command is created in working memory ("Imagine …") and additional procedural knowledge interprets that as a command to Soar's spatial-visual system (SVS) and sends it a symbol for the word "WOW." The visual representation is then retrieved from long-term perceptual memory and added to SVS's internal memory.

The next command for you to perform is to rotate the word, so it is upside down. In Soar, the command is once again parsed, with its meaning extracted through a combination of processing of procedural memory and semantic memory. Procedural knowledge then interprets that command and sends it to SVS. SVS uses long-term perceptual memory to identify that it is the word "MOM," which is added to working memory. Procedural memory then retrieves the internal meaning of that word from semantic memory.

The last command is for you to say the last time you saw the sister of that person. After language interpretation, procedural knowledge initiates a retrieval from semantic memory for the sister of your mother. Once information identifying your aunt is retrieved (if she exists and is known), procedural knowledge initiates a retrieval from episodic memory for the most recent time you saw her, and the results of that retrieval are used to generate a response (carried out by knowledge in procedural memory).

In Soar, agent behavior is driven by the interaction between the contents of working memory, which describe the agent's current goals, situation, and intermediate results of reasoning, and procedural memory, which encodes the agent's skills and processing knowledge. Other modules provide non-symbolic reasoning (SVS) and long-term declarative knowledge (semantic and episodic memory).

## 2. Deliberate Behavior: Selecting and Applying Operators

Soar organizes knowledge about conditional action and reasoning into *operators*. An operator can be an internal action, such as adding numbers together, retrieving information for episodic or semantic memory, or rotating an image in the spatial memory system. An operator can also be an external action such as initiating forward motion or turning in a mobile robot, producing a natural language statement, or accessing an external software system over the internet.

To support dynamic integration of knowledge, Soar decomposes the knowledge associated with an operator into three functions – proposing potential operators, evaluating proposed operators, and applying the operator. The use of operators contrasts with rule-based systems where the processing cycle involves selecting and firing a single rule, and where the knowledge for those functions is forever bound together as conditions (if-part) and actions (then-part). In contrast, the knowledge in Soar for each of those functions is represented as independent rules, which fire in parallel when they match the current situation. So in Soar, rules are not themselves alternative actions, but instead units of context-dependent knowledge that contribute to making a decision and taking action. Thus, there are rules to propose operators (operator proposal), evaluate proposed operators (operator evaluation), and apply the selected operator (operator application).

For example, consider a simple blocks world task, where the goal is to create a tower on a specific location, and there are operators for stacking and unstacking blocks. For the unstack operator, there is a proposal rule that proposes moving a clear block that is stacked on another block onto the table. For every such block, this rule would fire once, creating an associated operator. So if there are three stacked blocks, three unstack operators are created. There could be an operator evaluation rule that rejects a proposed operator that moves a block that is already part of the tower. Finally, there would be a rule that tests that an unstack operator is selected and creates a motor command for unstacking the block associated with that operator.

This example assumes that rules are available for every phase. As described in Section 3, when the available knowledge is insufficient to make a decision or apply an operator, an impasse arises in Soar, and the proposal, evaluation, and/or application phases are themselves carried out via recursive operator selections and applications in a substate., This impasse-driven process is the means through which complex, hierarchical operators and reflective meta-reasoning (including planning) are implemented in Soar, instead of through additional task-specific modules. Thus, an operator such as moving to the next room, setting the table, or deriving the semantics from natural language input, can be proposed and selected, but then implemented in a substate through more primitive operators in a substate.

Working memory contains the information that is tested by rules. It includes goals, data retrieved from long-term memories, information from perception, results of internal operators, and so on. Soar does not have any predefined structure for the contents of working memory except for the buffers that interface to other modules. Thus, there is no predefined structure for goals or even operators. The one exception is that to support the proposal and selection of operators, Soar has a special data structure called a *preference*. Preferences are created in the actions of rules and added to *preference memory* (see Figure 1). There are two classes of preferences. *Acceptable* preferences are created by operator proposal rules to indicate an operator is available for selection independent of whether it should be selected. Acceptable preferences are also added to working memory so that the evaluation rules can detect which operators have been proposed. Evaluative preferences, of which there are many types, specify information about whether an operator should be selected for the current situation. These are described in more detail below.

Figure 2 shows the decision cycle that implements this operator-centric approach to processing, with each phase described in more detail below.

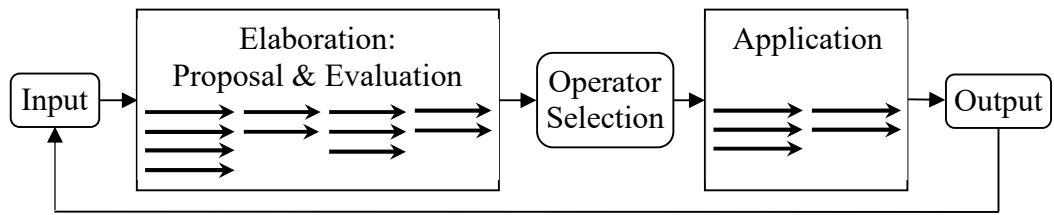

Figure 2: The decision cycle.

## 2.1 Input Phase
Starting at the left of Figure 2, the input phase processes data from perception, SVS, and retrievals from semantic and episodic memory, and adds that information to the associated buffers in working memory.

## 2.2 Elaboration Phase
The elaboration phase is where rules fire that elaborate the situation (not mentioned earlier), propose operators, and evaluate operators. All rules in Soar fire only once for a specific match to data in working memory (this is called an *instantiation*), and a given rule will fire multiple times if it matches different structures in working memory.

The rules that fire in this phase make *monotonic additions* to either working or preference memory, and the structures they create are valid only as long as the rule instantiation that created them matches. Thus, when changes to the working memory cause the rule instantiation to no longer match, the structures created by it (elaborations or preferences) are *retracted* and removed from their respective memories. For example, if a robot has a rule that proposes moving forward when there is no obstacle blocking its path, that rule will retract when an object gets within range, removing the proposal. This assertion/retraction process is an example of justification-based truth maintenance (Forbus & DeKleer, 1993).

There is no ordering of firing and retracting among the different types of rules – new firings and retractions occur in parallel. However, there is a common progression that starts with a wave of elaboration rule firings, followed by a wave of operator proposal, and finally a wave of operator evaluation. Different type rules that newly match or retract intermix within a wave. For example, a proposal rule that tests only data from input will fire at the same time as elaborations that test input. When no additional rules fire or retract, *quiescence* is reached, and control passes to operator selection.

Operator application rules, even those that newly match because of changes from input, *do not* fire during the elaboration phase. They fire only during the operator application phase. Given the dynamics of elaboration, this ensures that they fire only when their tested operator is preferred for the current situation.

### 2.2.1 Elaboration Rules
Elaboration rules create new structures entailed by existing structures in working memory. Elaborations can simplify the knowledge required for operator proposal, evaluation, and application by detecting useful regularities or abstractions. An example could be a rule that determines if an object is close enough to be picked up by testing that the object is not more than 8 inches away. By computing this once in a single rule, other rules can test the abstraction (close-enough-for-pickup) instead of the specific distance.

### 2.2.2 Operator Proposal Rules

In parallel with elaborations (or following them if dependent on them), *operator proposal* rules test the current situation to determine if an operator is relevant. If it is, they propose the operator by creating an acceptable preference and a representation of the operator in working memory. Essentially, they test the *preconditions* or *affordances* of an operator and create an explicit representation, which often will include parameters, such as the block to unstack. Often, task knowledge is incorporated into the proposal to avoid proposing all operators that are legal in a situation. For a mobile robot, an operator proposal rule might test that if the goal is to pick up an object, and the robot can see the object, is lined up with the object, is not close enough to the object (testing the absence of the result computed by the elaboration above), and there are no objects between it and the object, then it will propose moving forward.

### 2.2.3 Operator Evaluation Rules

Once operators have been proposed, *operator evaluation* rules test the proposed operators and other contents of working memory (such as the goal) and create preferences that determine which operator is selected. Soar has preferences that allow assertions of an operator's value relative to another operator (A is better than B, A is equal to B) or absolute value (A is as good as can be achieved, B is the worst that can be achieved, or C should be rejected). There are also numeric preferences that encode the expected future reward of an operator for the current situation as matched by the conditions of the rule. Section 5 describes how these rules and preferences are used with reinforcement learning. All preferences newly created are added to *preference* memory to be processed during the operator selection phase.

## 2.3 Operator Selection Phase

Once quiescence is reached, a fixed decision procedure processes the contents of preference memory to choose the current operator. If the available preferences are sufficient to make a choice, a structure is created in working memory indicating the selected operator. If the preferences are inconclusive or in conflict, then no operator is selected, and an impasse arises as described in Section 3.

## 2.4 Operator Application Phase

Once an operator is selected, operator application rules that match the selected operator fire to apply it. These rules make *non-monotonic* changes to working memory and do not retract those changes when they no longer match. Operator applications have multiple functional roles:
- Internal reasoning steps that modify the internal state of a problem. For example, if an agent is counting the number of objects in a room, and the count is maintained in working memory, a counting operator will increment the count.
- External actions that create commands in the motor buffer. For a robot, once the move forward operator is selected, a rule would fire and create the move forward command in the motor buffer.
- Mental imagery actions that create commands in the SVS buffer (see Section 8 for more details). These actions include projecting hypothetical objects, filtering information, or performing actions on SVS structures. For a robot, it could project the locations of objects that are obscured by its arm into SVS as the arm moves in front of its camera.
- A retrieval cue that is created in either of the semantic or episodic memory buffers (see Sections 6 and 7 for more details). For a robot, as it moves into a new room, it might retrieve the layout of the room from semantic memory. When asked to retrieve an object, it can query episodic memory to recall its last known location.

Operator application terminates when there are no more rules to fire or if the selected operator is no longer the best choice. To determine this second case, the decision procedure is rerun after each wave to ensure that the current operator should stay selected. The selected operator can change if either the rule that proposed it retracts or there are sufficient changes in other preferences such that another operator is preferred. If that is the case, the current operator is deselected, but a new operator is not selected until the next operator application phase. An impasse arises if the same operator stays selected in the next operator selection phase, which indicates there is not sufficient knowledge to immediately apply the operator, either because it takes time to apply in the environment, or there is not sufficient application knowledge.

**2.5 Output Phase**
Once the application phase completes, the output phase sends any structures created in buffers to their respective modules. After the output phase, processing returns to the input phase

## 3. Impasses and Substates: Responding to Incomplete Knowledge

In Section 2, we assumed that an agent had sufficient knowledge to propose relevant operators, evaluate and select a single operator from those proposed, and then apply the selected operator. Many AI systems make similar assumptions. If those assumptions are violated, AI systems either make random decisions or are unable to proceed. Soar embraces a philosophy that additional relevant knowledge can be obtained through additional deliberate reasoning and retrieval from other sources, including internal reasoning with procedural memory (e.g., planning), reasoning with non-symbolic knowledge, retrieval from episodic memory or semantic memory, or interaction with the outside world. This is essentially the philosophy of "going meta" when directly available knowledge is inadequate. Furthermore, Soar commits to a single approach for both deliberation and meta-reasoning, which differs from architectures that have separate meta-processing modules, such as MIDCA (Cox et al., 2016) and Clarion (Sun, 2016).

In Soar, if insufficient or conflicting knowledge is detected during operator selection, a decision cannot be made, and an *impasse* arises. There are three types of impasses that correspond to the different types of failures of the decision procedure that are related to the different types of knowledge described earlier: operator proposal, operator evaluation, and operator application. These types of impasses are listed below:
- State no-change: No operators are proposed. This usually indicates that new operators need to be created for the current situation.
- Operator tie/conflict: Multiple operators are proposed, but the evaluation preferences are insufficient or in conflict for making a decision. Usually, this involves determining which proposed operator should be selected and creating the appropriate preferences to make that happen.
- Operator no-change: The same operator stays selected across multiple decision cycles. This usually indicates that there is insufficient knowledge to apply an operator, but it can also arise when an operator is selected whose actions require multiple cycles to apply in the external environment. In that situation, the response is usually for the agent to wait for the action to complete.

To localize reasoning about impasses from the original task information, Soar organizes information in working memory into *states*. The initial contents of working memory are all linked to a single state, called the *topstate*. When there is an impasse, a new state is created, (a *substate*[3]) in working memory that links to the existing state (its *superstate*). The substate has its own preference memory and the ability to select and apply operators without disrupting any of the processing in superstates. Through the link to the superstate, it has access to all superstate structures. It has buffers for semantic memory and episodic memory, but not perception and motor, which must go through the topstate so that all interaction with the outside world can be monitored by reasoning in the topstate.

---
[3] Throughout Soar's history, we have used both substate and subgoal as names for this structure. I use substate because it is technically more correct. Any subgoal information is a subset of the structures in a state.

Procedural memory matches against structures in the substate just as in the topstate. Operators can be proposed, selected, and applied using the same decision procedure, but applied to the preferences that are created locally in the substate. If there is insufficient knowledge in the substate to select or apply an operator, another impasse arises, which leads to a stack of states. There is a tradeoff in that deliberate reasoning in a substate takes more time. Execution time can be monitored in the substate so that a non-optimal (or even random) decision can be made if time is an issue.

When an operator in a substate creates and modifies structures in a superstate, those changes are the *results* of the substate. If the result is an elaboration in the superstate (state elaboration, operator proposal, or operator evaluation), it persists only as long as the superstate structures responsible for its creation exist. If the result is part of the application of an operator selected in the superstate (a *superoperator*), such as creating a motor command, the result does not retract.

The decision cycle remains active across all states, and an impasse is resolved whenever a new decision is possible in a superstate, either through the creation of new preferences or through changes in the state (including changes in perception) that then lead to changes in preferences in the substate. When an impasse is resolved, the substate terminates and all non-result substate structures are automatically removed. Through this approach, Soar agents remain reactive to relevant changes in their environment, even when there are many active substates. For example, if an agent is attempting to build a tower and has an operator proposed for multiple blocks, there could be an impasse to choose from among those operators. If a human removes all the blocks but one, the resulting changes to input will lead to rule retractions so that only a single operator is proposed, and a decision will be made, resolving the impasse.

### 3.1 Deliberate Operator Application: Hierarchical Task Decomposition

For many problems, it is possible to define operators that apply immediately through the application of rules, so that all decision-making is in terms of primitive actions, such as moving forward or turning in a mobile robot. However, it is frequently useful to reason with more abstract operators that require dynamic combinations of primitive operators and other abstract operators. For example, if we want our robot to clean a room, it would be useful to have operators such as fetch, throw-away, or store. These operators are not primitive and require the execution of multiple low-level operators, which themselves may require problem solving to perform, leading to *hierarchical task decompositions*. Hierarchical task decomposition arises naturally in Soar when an abstract operator is selected, such as fetch, and there are no rules that directly implement it. This leads to an impasse and the creation of a substate. As part of this approach, there must be procedural knowledge that is sensitive to the existence of such an impasse, which then proposes operators for implementing fetch, such as remember-location, find-object, go-to-object, pickup-object, and return-to-location, and selecting between them. Some of these operators are primitive, such as remember-location, while the others are abstract and lead to substates with additional problem solving, ultimately bottoming out in movement and manipulation operators. Figure 3 shows a three-level hierarchy of operators (but not substates) where the application of a move-block operator is decomposed into pick-up and put-down. Pick-up and put-down are implemented in even more primitive operators that direct control the gripper.

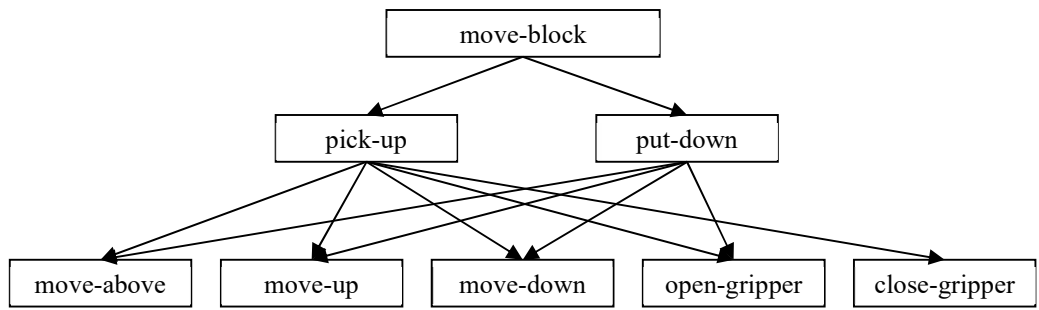

Figure 3: Hierarchy of operators for move-block.

Soar's approach to hierarchical decomposition has similarities to those used in Hierarchical Task Networks, Hierarchical Goal Networks, or options in RL. A few distinguishing characteristics of Soar's approach are that the hierarchical decomposition is determined by the knowledge available at run time. There is no fixed declarative structure that specifies the decomposition, nor are there fixed sequences of sub-operators that are applied. Within a substate, all of Soar's processing is available, which includes planning, retrievals from semantic and episodic memory, and interaction with the external environment.

### 3.2 Deliberate Operator Selection

When the preferences are insufficient to pick a single operator, an impasse arises. The purpose of the resulting substate is to determine which operator should be selected and to create additional preferences so a decision can be made.[4] This is a case when an automatic parallel process (rule firings) is insufficient, it is supplemented by a deliberate, sequential process (operator selection and application) that can explicitly reason about each proposed operator and how it compares to the other proposed operators. This naturally leads to look-ahead planning and other deliberate approaches to operator evaluation. Consider the example of where the robot wants to move to another room and has abstract operators for moving to an adjacent room. If there are no evaluation rules, an impasse arises and in the substate, the agent can use internal models of its operators to imagine the situation after applying these different operators. It could then evaluate those situations based on how close they are to the desired room or keep planning until a path to the desired room is found. Different look-ahead search methods, such as iterative-deepening approaches to best-first search or alpha-beta, arise depending on the knowledge available for evaluating states and combining the evaluations (Laird et al., 1986).

### 3.3 Summary of Impasses and Substates

Impasse-driven substates allow an agent to automatically transition from using parallel procedural knowledge to using more reflective, metacognitive reasoning when the procedural knowledge is insufficient to select and/or apply operators (Rosenbloom, Laird & Newell, 1986). It provides a means to incorporate any and all types of reasoning that are possible in Soar in service of the core aspects of proposing, selecting, and applying operators, so that as described below, there is a portal for incorporating additional knowledge and unconstrained reasoning, including deliberate access to semantic and episodic memories, non-symbolic reasoning, planning, reasoning about others (Laird, 2001), etc.

## 4. Chunking: Learning New Rules

Impasses arise when there is a lack of knowledge to select or apply an operator. The processing in a substate creates knowledge to resolve the impasse, creating an opportunity for learning. In Soar, c*hunking* compiles the processing in a substate into rules that create the substate results, eliminating future impasses and substate processing. Thus, chunking is a learning mechanism that converts deliberate, sequential reasoning into parallel rule firings.

---

[4] The preference scheme is such that additional preferences can always eliminate an impasse that arises for operator selection. In the extreme, this requires creating new copies of some of the tied operators.

Chunking is automatic and is invoked whenever a result is created in a substate. It analyzes a historical trace of the processing in the substate, determining which structures in the superstate had to exist for the results of the substate processing to be created. Those structures become the conditions of a rule, and the results become the actions. Independent results in a substate lead to the learning of multiple rules.

In more detail, when results are created, Soar *back-traces* through the rule that created them, finding the working memory elements that were tested in its conditions. All of those working memory elements that are part of a superstate are saved to become conditions along with additional tests in the conditions of the associated rules, such as equality and inequality tests of variables used in rule conditions, or comparative tests of numbers, and so on. Tested working memory elements that were local to the substate lead to further back-tracing, which recurs until all potential conditions are determined. Once back-tracing is complete, chunking creates a rule from the saved structures.

Chunking has recently been completely reimplemented and the new implementation is based on recent analyses and a subsequent design that ensures that the resulting rules are correct relative to the reasoning in the substate and as general as possible without being over general. This new approach is called explanation-based behavior summarization (EBBS; Assanie, 2022).

Chunking can learn all the types of rules encoded in procedural memory and what it learns is dependent only on the knowledge used in the substates to create results. The breadth of that processing means that chunking can learn a surprisingly wide variety of types of knowledge. Steier et al., (1987) and Rosenbloom et al., (1993) are collections of early examples of the diversity of learning possible with chunking.
- **Elaboration:** In any impasse, if the processing in the substate creates monotonic entailments of the superstate, Soar learns elaboration rules.
- **Operator Proposal:** When there is a state no-change impasse, it is resolved through the creation of acceptable preferences for new operators, which lead to the learning of operator proposal rules.
- **Operator Evaluation:** When there is a tie or conflict impasse, and a substate generates new preferences, chunking learns operator evaluation rules. When the substate processing involves planning, this involves compiling planning into knowledge that is variously called search-control, heuristics, or value functions. When those are numeric preferences, the rules created by chunking will be RL rules that are initialized with whatever evaluation was generated. In the future, those rules can be further tuned by reinforcement learning (Laird et al., 2011).
- **Operator Application:** For operator no-change impasses where the substate implements an operator through hierarchical decomposition, interpreting declarative representations from semantic memory, or recalling past examples from episodic memory, chunking learns operator application rules (Laird et al., 2010). For operators used to simulate external behavior, this corresponds to model learning.

One limitation is that chunking requires that substate decisions be deterministic so that they will always create the same result. Therefore, chunking is not used when decisions are made using numeric preferences. We have plans to modify chunking so that such chunks are added to procedural memory when there is sufficient accumulated experience to ensure that they have a high probability of being correct.

A potential concern with chunking is that the costs of the analyses it performs to create a chunk could surpass its benefits. Our empirical results show that such overhead is minimal and that the performance improvements are much greater than the costs (Assanie, 2022).

## 5. Reinforcement Learning: Learning Control Knowledge from Reward

Given the model of deliberation as the selection and application of operators, there is an opportunity to tune the selection of operators as the agent receives feedback on its activities, such as from the achievement of a goal, a failure, or some other kind of internal or external reward. The general formulation of this problem is called Reinforcement Learning (RL; Sutton & Barto, 2018) and has a long history in AI. RL modifies selection knowledge so that an agent's operator selections maximize future reward.

Using RL in Soar is straightforward. The first step is to create operator evaluation rules that create numeric preferences, aptly called *RL rules* (Nason & Laird, 2005). The conditions of an RL rule determine which states and proposed operators the rule applies to, and the numeric value of the preference encodes the expected reward (Q value) for those states and operators. After an operator applies, all RL rules that created numeric preferences for it are updated based on the reward associated with the state and the expected future reward. The expected future reward is the sum of the numeric preferences for the next selected operator.[5] Even when no reward is created for a state, the values associated with operators proposed for that state propagate the expected reward back to RL rules used in the previous decision.

RL influences operator selection only when the other, non-RL preferences are insufficient for making a decision. Thus, RL rules do not have to be relied on to avoid dangerous situations (or directly seek advantageous situations) when there are other sources of knowledge (such as other procedural knowledge or advice from a human) readily available. Then RL can be used only for those situations where there is uncertainty. This hybrid approach can lead to much faster learning using RL, as well as less risk of dangerous behavior.

An example use of RL rules with our robot might be a set of rules for moving and turning in relation to an object the robot wants to pick up. Each rule would test different distances and orientations of the object relative to the robot, and a specific operator for either moving forward fast or slow, or turning left or right. The action would be a numeric preference (Q value) for that operator. Assuming a reward that decreases over time, with experience, the rules will learn to prefer the operators that are best for quickly achieving the objective. As mentioned above, the RL rules can be supplemented with evaluation rules that always avoid collisions from the start, which are usually easy to include when the agent is being developed.

Reward is created by rules that test state features and create a reward structure on the state. In a simple case, a rule detects that a goal has been achieved and creates a positive reward, such as when the object is picked up. However, rules can also compute reward by evaluating intermediate states or converting sensory information (including external reward signals) into a representation of reward on the state. In our example, an intermediate reward could be proportional to the distance to the object, although that could have some issues in this case. Soar currently has no pre-existing intrinsic reward; however, intrinsic reward based on appraisal theory has been used in an experimental version of Soar (Marinier et al., 2009).

---

[5] Soar supports both Q-learning and SARSA, as well as eligibility traces. It has parameters for learning rates and discount rates, which are fixed at agent initialization.

All created rewards are summed to give the total reward for the state. As operators are applied, reward changes as rules that create reward values fire (or retract) in response to changes in the state.

One intriguing aspect of RL in Soar is that the mapping from state and operator to expected reward (the value-function) is represented as collections of relational rules. When there are multiple rules that test different state and operator features, they can provide flexible and complex value functions that map from states and operators to an expected value, supporting tile coding, hierarchical tile coding, coarse coding, and other combination mappings, which can greatly speed up learning (Wang & Laird, 2007). Furthermore, RL rules can be learned by chunking, where the initial value is initialized by the processing in a substate, and then subsequently tuned by the agent's experience and reward.

Finally, RL in Soar applies to every active substate, with independent rewards and updates for RL rules across the substates. Thus, Soar naturally supports hierarchical reinforcement learning for all different types of problem solving and reasoning, including when planning is used in substates (model-based RL), or in the topstate (model-free RL).

## 6. Semantic Memory

Soar has two additional long-term symbolic declarative memories in addition to procedural memory. Semantic memory (Wang & Laird, 2006; Derbinsky et al., 2010, Jones et al., 2016) encodes facts that an agent "knows" about itself and the world, while episodic memory (described below in Section 7) encodes what it "remembers" about its experiences. Thus, semantic memory serves as a knowledge base that encodes general context-independent world knowledge, but also specific knowledge about an agent's environment, capabilities, and long-term goals.

Semantic and episodic memories differ from procedural memory in how knowledge is encoded (as graph structures instead of rules), how knowledge is accessed (through deliberate cues biased by metadata as opposed to the matching of working memory to rule conditions), what is retrieved (a declarative representation of the concept or episode that best matches the cue versus the actions of a rule), and how they are learned (described below). A common question is why the information in semantic memory cannot be maintained in working memory. Unfortunately, the cost of matching procedural knowledge against working memory increases significantly with working memory size, making it necessary to store long-term knowledge separately.

Concepts in semantic memory are encoded in the same symbolic graph structures as used in working memory.[6] As noted above, knowledge is retrieved from semantic memory by the creation of a cue in the semantic memory buffer. The cue is a partial specification of the concept to be retrieved, such as a room that has a trash can in it. The cue is used to search semantic memory for the concept that best matches the cue, and once that is determined, the complete concept is retrieved into the working memory buffer, where it can then be tested by procedural memory to influence behavior. Thus, semantic memory provides flexibility in what aspects of a concept are used as a cue to retrieve it. The same trashcan concept might also be retrieved using its color, its brand, or its size (assuming each of those is a defining characteristic).

Soar uses a combination of base-level activation and spreading activation to determine the best match, as used originally in ACT-R (Anderson et al., 2004). Base-level activation biases the retrieval using recency and frequency of previous accesses of concepts (and noise), whereas spreading activation biases the retrieval to concepts that are linked to other concepts in working memory. Together, these model human access to long-term semantic memory (Schatz et al., 2018), and have been shown to retrieve the concept most likely to be relevant to the current context (as defined by working memory). For example, if a retrieval is attempted for the concept associated with the concept "bank," and the agent has recently had the definition associated with a financial institution in working memory, that concept would get an activation boost from recency. In another situation, where someone is canoeing down the river, the concept of river would spread activation to the concept related to the bank of a river, improving the likelihood of its retrieval.

---

[6] Semantic (and episodic) memory can include symbols that are pointers to modality-specific structures (such as images), and those structures can then be retrieved in Soar's modality-specific memory (SVS described below).

Semantic memory can be initialized with knowledge from existing curated knowledge bases (such as WordNet or DBpedia) and/or built up incrementally by the agent during its operations. As of yet, Soar does not have an automatic learning mechanism for semantic memory, but an agent can deliberately store information at any time. In our robot, semantic memory would hold the agent's map of its environment, either preloaded from some other source or dynamically constructed as the robot explores its world. Other common uses are to maintain representations of words and their meaning for language processing, knowledge about different agents the robot interacts with, and declarative representations of hierarchical task structures it learns from instruction.

## 7. Episodic Memory

In contrast to semantic memory, which contains knowledge that is independent of when it was learned, Soar's episodic memory (Nuxoll & Laird, 2012; Derbinsky et al., 2012; Jones & Laird, 2019) contains memories of what has been experienced over time (Tulving, 1983). An episode is a snapshot of the structures in the topstate. A new episode is automatically stored at the end of each decision. Episodic memory provides an agent with the ability to remember the context of past experiences as well as the temporal relationships between experiences. For our robot, this provides a memory of which rooms it has recently visited, where it has seen people, what interactions it has had with them, and so on.

Similar to retrievals in semantic memory, and for similar reasons, retrievals from episodic memory are initiated via a cue created in the episodic memory buffer by procedural knowledge. Unlike semantic memory, a cue for an episodic retrieval is a partial specification of a complete state, as opposed to a single concept. Episodic memory is searched, and the best match is retrieved (biased by recency) and recreated in the buffer. Once a memory is retrieved, memories before or after that episode can also be retrieved, providing the ability to replay an experience as a sequence of retrieved episodes or to move backward through an experience to determine factors that influenced the current situation, including prior operators, but also changes from the dynamics of the environment.

Soar minimizes the memory overhead of episodic memory by storing only the changes between episodes and uses indexing to minimize retrieval costs. However, memory does grow over time, and the cost to retrieve old episodes slowly increases as the number of episodes grows, whereas the time to retrieve recent episodes remains constant. An agent can further limit the costs of retrievals by explicitly controlling which aspects of the state are stored, usually ignoring frequently changing low-level sensory data.

Episodic learning has often been ignored as a learning method because it does not have generalization mechanisms. However, even without generalization, it supports many important capabilities:

- Virtual sensing of remembered locations and objects that are outside of immediate perception (Nuxoll & Laird, 2012).
- Learning action models for internal stimulation from memories of the effects of past operators that have external actions (Laird et al., 2010; Jones, 2022).
- Learning operator evaluation knowledge via retrospective analysis of previous behavior (Mohan & Laird, 2013).
- Using prospective memory to trigger future behavior, by imagining hypothetical future situations that are stored in memory and then recalled at the appropriate time (Li & Laird, 2015).
- Using a string of episodes to reconstruct and debug a particular course of action.

## 8. Spatial-Visual System

All the representations in Soar's short and long-term memories described so far are symbolic, with non-symbolic metadata used for biasing decision making and retrievals from long-term memories. For real-world environments, many of the symbol structures in working memory must be grounded in the perception of objects in the world. Furthermore, there are forms of reasoning that can be much more efficient with modality-specific representations, such as mental imagery (Kosslyn et al., 2006). Purely symbolic representations abstract away critical details and do not efficiently support non-symbolic data processing, such as simulating movement and interactions of objects in 2D or 3D space.

In Soar, perceptual grounding and modality-specific non-symbolic representations and processing are supported by the spatial-visual system (SVS; Lathrop & Laird, 2007; Lathrop et al., 2011; Wintermute, 2010; Mininger & Laird, 2019). SVS is an intermediary between symbolic working memory and the non-symbolic components of a Soar agent, including perception and motor control, and long-term modality-specific memories. As described below, information flows not only bottom-up from perception through SVS into working memory but also flows top-down from working memory to SVS to support reasoning over hypothetical non-symbolic representations.

Soar's approach to integrated symbolic and modality-specific representations contrasts with architectures that have unified representations where both symbolic and non-symbolic representations are combined in a single memory. The separation in Soar allows for independently specialized and optimized processing of each type of representation but requires an interface for transferring information between them.

SVS is currently restricted to visual input, which includes 2D image-based/depictive representations for video games and 3D representations of objects with associated spatial metric information (scene graphs) for real-world robots and 3D simulation environments. To control the flow of information from SVS to working memory, an agent uses operators to issue commands to SVS that create *filters*. Filters automatically extract symbolic properties (object `o45` has `color g34`, has `shape box`, has `image i45`, has `height 34 inches`, …) and relations (object `o45` is to the `left-of` object `o65`).

SVS supports hypothetical reasoning over spatial-visual representations through the ability to *project* non-symbolic structures into SVS, such as projecting a ball into a scene. An agent can perform modality-specific operations on the objects in SVS (real and imagined), such as rotation, scaling, and motion, and detect relations such as collisions, to determine the outcome of potential motor actions in the current state. For example, a Soar agent can simulate the motion of a vehicle along a path to determine if it will collide with other objects in planning (Wintermute, 2010). Thus, SVS mediates interactions between motor actions and perception in 3D robotic environments.

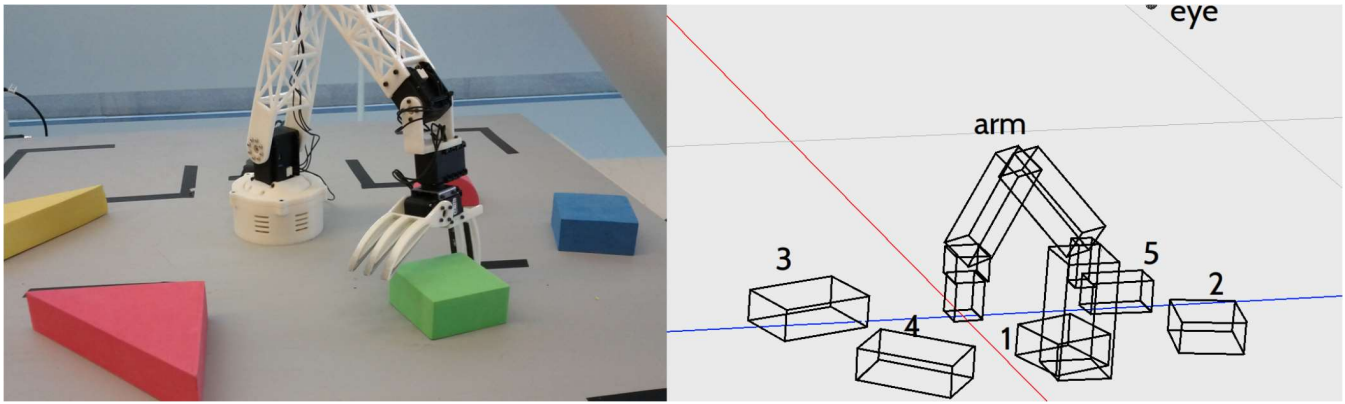

Figure 4: An example of the tabletop arm (left) and the corresponding spatial information encoded in SVS derived from the perceptual system (right). From Mininger (2019).

Figure 4 shows an example of SVS's use while planning to move a robot arm. A representation of the arm together with objects in the environment is maintained in SVS and in parallel, a symbolic representation is maintained in working memory. When the arm is moved in such a way that it obscures an object (such as object 5, which is red and barely visible in the image), the agent uses its knowledge of 3D space and object persistence to understand that the obscured object has not disappeared, and it can maintain a symbolic representation of the object in working memory (Mininger, 2019).

## 9. Summary and Review

In this section, I first summarize the different levels of reactive, deliberative, and meta processing in Soar. Next is a discussion of the different types of learning, followed by a list of some unusual combinations of reasoning and learning that are possible in Soar.

### 9.1 Levels of processing

In describing Soar, it is useful to think in terms of different levels of processing and times scales as originally defined by Newell (1990), as shown in Figure 5. Those time scales do not directly apply to Soar's real-time processing, which can be hundreds to thousands of times faster, however they provide a mapping to human performance.

| Soar Level | Processing in Soar | Newell Level |
|---|---|---|
| Substates | Processing in a substate | Operations, unit task, and above > ~1 sec |
| Decision Cycle | Sequential selection and application of an operator | Deliberate Act ~100 ms |
| Module | Parallel processing within a module Rule matching and memory retrievals | Neural Circuit 10ms |
| Architecture | C code Module processing | Neuron ~1 ms |

Figure 5: Levels of processing and representations in Soar aligned with Newell's bands.

Starting at the bottom of Figure 5 is the architecture implemented in computer code, which in Soar's case is C/C++. An agent has no access to this level. Above this level, the processing of Soar is in terms of individual modules, such as the processing in procedural memory to match rules against working memory and the retrievals from semantic or episodic memory. Even though Soar is implemented on serial computers, conceptually the individual modules run in parallel whenever they have input to process.

The decision cycle is the next level, which imposes sequential behavior via operator selection and application. This corresponds to Newell's first cognitive level of deliberate acts. For impasse-free agents, this is their "top" level, with all processing happening here or below.

The next level is for the processing in substates in response to impasses. Although the processing in a substate uses the same decision cycle, conceptually this is metalevel processing: deliberation to make a decision, apply an operator, or create a new operator. Thus, in place of what would otherwise be a single decision if there were no impasse, in the substate, the same conceptual function (operator proposal, evaluation, application) is carried out by the selection and application of operators across multiple decisions. Those decisions can have impasses/substates as well. Substate operators can perform processing that is not possible in a single decision, such as making multiple retrievals of data from episodic or semantic memory, simulating hypothetical states in SVS, or even interacting with a human. This level is where full-scale sequential reflective reasoning can be used when the fast, automatic retrieval from procedural memory is insufficient for making a decision. Thus, this is Newell's second cognitive level and can extend further up his hierarchy (not included here) depending on the available knowledge.

One important feature is that the processing levels for a task are not static over time. Chunking automatically converts processing in substates into rules, so that as an agent gets experience with a task, the higher-level processing levels are replaced by rules that select and apply operators.

It is natural to attempt to map these levels onto Kahneman's (2011) System 1 (instinctive and emotional) and System 2 (deliberative and more logical). One possible mapping is that architectural processing (including low-level perception, retrievals from memory) is at System 1, and the processing in substates is at System 2. Interestingly, many types of processing described by Kahneman as included in System 1 require multiple individual decisions and multiple accesses to long-term declarative memory, such as adding two numbers or reading a text on a billboard. Thus, another possibility is that System 1 includes processing where there are no substates, but this mapping seems overly simplistic. System 1 and System 2 as conceptual categories initially seem well defined, but once you try to map them onto specific computational processes, there are subtleties and questions that require more deliberate analysis.

## 9.2 Varieties of Learning

Figure 6 is a summary of Soar memory and learning systems. The image memory system is still experimental, although variants have been used on past systems that relied on SVS. What this figure does not show are the variety of contents that each learning mechanism stores. For example, chunking can learn operator proposal, operator evaluation, or operator implementation knowledge, and the exact content depends on the reasoning performed in the substate that generated the result.

| Memory/Learning System | Source of Knowledge | Representation of Learned Knowledge | Retrieval of Knowledge |
|---|---|---|---|
| Chunking | Traces of rule firings in subgoals | Rules | Exact match of rule conditions |
| Semantic Memory | Existence in Working memory | Graphs | Partial match biased by activation |
| Episodic Memory | Working memory co-occurrence | Episodes: Snapshots of working memory | Partial match, temporally adjacent episodes |
| Reinforcement Learning | Reward and preferences | Rules that create numeric preferences | Exact match of rule conditions |
| Image Memory | Image short-term memory | Image | Deliberate recall using symbolic referent |

Figure 6: Summary of Soar's Memory and Learning Systems.

### 9.3 Combinations of Reasoning and Learning

One unique aspect of cognitive architectures is that they can combine multiple approaches to reasoning and learning depending on the available knowledge. Below are examples that have arisen in Soar agents.

1. Reinforcement learning is used to learn retrievals from episodic memory (Gorski & Laird, 2011).
2. Mental imagery is used to simulate actions (in a video game) and detect collisions that inform reinforcement learning (Wintermute, 2010).
3. Chunking is used to learn RL rules whose initial values are derived from internal probabilistic reasoning that estimate the expected value. Those values are then tuned by experience (Laird, 2011).
4. Substates are used to reason about the actions of another agent via a combination of self-knowledge (the simulation approach to Theory of Mind) and knowledge learned about the agent (Laird, 2001).
5. A variety of knowledge sources are used to model the action of operators for planning, including procedural memory, semantic memory, episodic memory, and mental imagery (Laird et al., 2010), all of which are compiled into rules by chunking.
6. Episodic memory, metareasoning, and chunking are combined in retrospective analysis that provides one-shot learning of new operator proposal and evaluation knowledge (Mohan, 2015).
7. Emotional appraisals are used as intrinsic reward for reinforcement learning (Marinier et al., 2009).
8. Episodic memory is used to store goals for future situations in prospective tasks (Li, 2016).

## 10. Evaluation of Soar as a General Cognitive Architecture

Soar is designed to support AI agents whose cognitive characteristics and capabilities approach those found in humans (Laird, 2012; Newell, 1990). Below I consider capabilities that I consider important for such agents. The first 9 are drawn (in order) from Newell (1990), Figure 1-7. Newell labels these as "constraints that shape mind." Newell's constraints 8, and 11-13 are not included as they apply mainly to biological systems. The others are drawn from a variety of sources. These vary between being constraints on the function/behavior of an agent and its structure. The ones concerning structure seem less fundamental as they constrain how the architecture is realized as opposed to what it does. For the structural ones, I attempt to include the functional capabilities engendered by the structure.

I rate Soar's capabilities as follows: "yes" means that Soar agents achieve this capability to a significant extent (although it is possible to deliberately create Soar agents that do not). "Partial" means that Soar agents have been demonstrated to achieve aspects of this capability, but it is not commonly available. "No" means there are no Soar agents that have demonstrated this capability. Those that are "partial" and "no" help define the future research agenda for Soar. Following each of my ratings is Newell's rating from 1990, which were "yes" or "no." We mostly agree, but there are cases where I believe Soar has demonstrated capabilities since 1990 that warrant at least a "partial" on my scale instead of a "no." I also am a bit stricter about when learning capabilities achieve a "yes." For an evaluation of ACT-R and connectionist systems on constraints 1-7, 9-10, see Anderson & Lebiere (2003).

1. **Behave flexibly as a function of the environment (yes, yes).** Behavior cannot be purely preplanned but must quickly and flexibly respond to changes in the environment. The levels of processing in Section 9.1 demonstrate how Soar supports complex, real-time goal-oriented behavior that is responsive to the current state of the environment.
2. **Exhibit adaptive (rational, goal-oriented) behavior (yes, yes).** Tasks require combinations of achieving specific situations (goals) and performing actions (such as in cooking). Tasks can have hierarchical structures where subtasks need to be achieved in pursuit of an overall task. Soar supports representing multiple concurrent complex tasks, including hierarchical tasks. It also supports tasks that are defined by goals and constraints (many games) and those that are defined in terms of executing a procedure (baking a cake). Moreover, Rosie, an agent developed in Soar, learns these different types of tasks from real-time natural language interaction with an instructor.
3. **Operate in real time (yes, yes).** An agent must be responsive to the dynamics of its environment, which for Soar is determined by the loop from perception, decision, to motor action. Empirical evidence is that a decision cycle time of around 50 msec. is required for real-time behavior. Soar achieves that even with large long-term memories (millions of items).
4. **Operate in a rich, complex, detailed environment (yes, yes).** Agents are embodied within an external environment and interact with it through sensors and effectors, and where the environment has its own dynamics that influence the agent's behavior and its ability to achieve its goals. Soar agents have been developed for over 20 real-world robots, a variety of high-fidelity simulations, and multiple digital environments.
5. **Symbol representations and reasoning (yes, yes).** Symbolic structures support the creation of composable representations that abstract away from details of perception, making reasoning more efficient, more general, open to introspection, as well as supporting communication with other agents. Soar supports deliberate reasoning over symbolic representations.
6. **Use language, both natural and artificial (partial, no).** Language is critical for communicating with other agents and in many theories is closely related to using symbolic representations. Over the years there have been multiple agents, such as NL-Soar (Lewis, 1993), that use artificial languages or restricted natural language, with Rosie being the best example (Lindes, 2022). However, this is not a capability that is available to any agent developed in Soar, and even Rosie cannot process unrestricted language.
7. **Learn from the environment and experience (partial, yes).** An agent needs to continually learn from its experience in its environment, acquiring new concepts and relations, improving its decision making, reducing the time it takes to make decisions, and building up a historical record of its experiences that it can use later for deliberate retrospection. It also needs to be able to learn from other agents. As described in Section 9.2, Soar has multiple general online incremental learning mechanisms that have been used in a wide variety of agents. Rosie uses them to acquire new tasks as well as to improve its performance on those tasks through practice. These mechanisms include procedure composition (chunking), episodic learning, and reinforcement learning. Soar does not have innate capabilities for learning perceptual categories. I rate this a "partial" because although learning is available to all agents, agents must be designed with learning in mind, and agents do not just learn to do new tasks from scratch without any of the pre-existing procedural and semantic knowledge included in Rosie. Further, there are still types of architectural learning that are missing, such as semantic learning, generalized perceptual category learning, and learning expectations or predictions of its own actions or the dynamics of its environment.

9. **Operate autonomously, but within a social community (partial, no).** Agents have a singular ongoing existence in their environment. Soar agents are completely autonomous, without direct human control. Any control is indirect through communication channels that feed in as input into working memory such as natural language. Some agents have run uninterrupted for 30 days. In addition, Rosie does exist in a (very limited) social community with its instructor. Other agents have been developed that operate in teams (Tambe, et al. 1995; Jones, et al. 1999). Thus, the evaluation of Soar is mixed, with strength in autonomy, but only limited examples of participating in a social community. However, it is not clear whether what is missing is some aspect of architecture as opposed to agent knowledge.
10. **Be self-aware and have a sense of self (no, no).** In order to fully participate in a community, as well as to have full meta-level capabilities, an agent must have some internal representation of itself and its own capabilities. In Soar agents, there is no explicit representation of the agent's "self." Many agents indirectly have access to some representation of their own capabilities through combinations of long-term procedural, semantic, and episodic memory; however, it is not explicitly available for reasoning beyond the ability to determine what it will do (or did) in a given situation.

Below are additional capabilities related to those in Newell's original list.
11. **Use modality-specific representations and reasoning (yes).** Modality-specific representations support efficient non-symbolic reasoning of sensor data and hypothetical situations that are difficult to represent and reason about using only symbolic structures. Through SVS, Soar supports representations of 2D representations of images and 3D representations of space and objects. Images and objects can be projected into SVS, and a variety of non-symbolic operations are supported for simulating behavior in an external environment.
12. **Use diverse types and levels of knowledge (yes).** An agent needs knowledge about how to perform tasks, about the objects and agents in its world, its experiences, etc. It must handle situations where its knowledge is incomplete and where reasoning or exploration are required to fill in the gaps, but it must also have the capacity to encode, store, and use vast bodies of knowledge. Another dimension of variability is the generality of its knowledge, where some knowledge applies to a range of tasks, whereas others can be specific to the details of a single task. Soar supports symbolic procedural, semantic, and episodic knowledge, as well as image/vision-based representations. It also supports different levels of knowledge. At one extreme, when knowledge of a task is complete, the decision cycle executes without interruption. When knowledge is incomplete, impasses lead to substates in which deliberate processing attempts to retrieve, generate, or discover the missing knowledge. In terms of the raw amount of knowledge, procedural, semantic, and episodic memories have all held not just thousands, but millions and even tens of millions of rules, facts, and episodes, while still maintaining real-time reactivity.
13. **Core and commonsense knowledge (no).** Although Soar supports the encoding and use of a wide variety of knowledge, we have not pre-encoded large bodies of innate knowledge that are available in every task. Thus, Soar agents have neither core knowledge systems (Kinzler & Spelke, 2007) nor large bodies of commonsense knowledge (Davis & Marcus, 2015). It is an open question as to whether this is an issue with missing innate knowledge, or whether it is an architectural issue in that Soar's procedural and semantic memories are insufficient for representing and accessing these types of knowledge and its learning mechanisms are insufficient for learning them. It is also possible that it is a combination of both.
14. **Reason about the past and the future (yes).** Beyond making decisions using information about the current situation, an agent needs the ability to plan and prepare for the future and have access to previous experiences to notice patterns and retrospective analyze its mistakes and failures. Episodic memory directly supports accessing past experiences for retrospective and even prospective reasoning.
15. **Use metacognition (partial).** Metacognition is the ability of an agent to reason about its own reasoning. This is related to having a sense of self (#10) but also includes reasoning about capabilities and knowledge. Soar's impasses and substates support recursive metacognition that has been used extensively for planning, retrospective and prospective reasoning, and predicting the behavior of other agents. However, the inherent capability is targeted to reasoning about action. It requires additional knowledge that is not ubiquitous in Soar agents for reasoning about other capabilities such as its own knowledge.
16. **Use emotion (partial).** The exact cognitive capabilities supported by emotion are still to be determined, and it might be that artificial agents do not need "real" emotion but only need to fake it for interacting with humans. However, emotion does have the potential of providing task general mechanisms for internal reward as defined by appraisal theory, as well as, characterizing situations that must be coped with. We have done some experimentation of using appraisal theory as a basis for intrinsic reward (Marinier et al., 2009), but more work needs to be done.

Other capabilities could be included in this list, such as being creative, using analogies, and so on. For now, our assumption is that these capabilities (and others) are implemented through knowledge and not through specialized support beyond what is already in the architecture. Natural language processing is an example where Soar agents use knowledge to implement natural language capabilities without direct architectural support. The architectural realization of other capabilities, such as additional forms of learning, the representation and use of commonsense knowledge and reasoning, and emotion processing, is more uncertain and these are definitely areas for future work.

In reviewing these 16 items, I have rated 8 as "yes," 5 as "partial," and 2 as "no." An optimist would take this progress as being well past the halfway point; however, even I have to admit that in working with Soar and the agents we build it in for 40 years, it feels like there are essential elements that are still missing. Maybe these elements are the "no's" (sense of self, and core and commonsense knowledge), or the missing parts of the "partials." If I try to dig a bit deeper, I come up with two theories.
1. Many of the items listed above involve evaluations of whether a capability exists, such as does the architecture has multiple learning mechanisms, does it support a variety of knowledge representations, and so on. Unfortunately, providing the existence of a capability does not ensure it is sufficiently broad to support the breadth of types of knowledge and reasoning required for a truly general autonomous agent. We tend to take examples of the existence of a capability for multiple tasks with different types of knowledge as proof of the universality of the full capability across all tasks an agent might encounter. Thus, a harsher evaluation would be that the most we have achieved for many of the items is "partial."
2. What I feel is most missing from Soar is its ability to "bootstrap" itself up from the architecture and a set of innate knowledge into being a fully capable agent across a breadth of tasks. Our agents do well when restricted to specific well-defined tasks. We have gone beyond that with Rosie, where we achieved success with its task learning capabilities. But without a human to closely guide it, Rosie is unable to get very far on its own, especially in being unable to learn new abstract symbolic concepts on its own, and similarly being unable to independently learn new abstract operators (although the IBMEA project briefly described in the appendix is making progress on this).

Thus, I see we see progress in cognitive architecture research, but still large gaps in our ability to create general autonomous agents that can marshal their learning on new tasks without heavy human supervision.

## Acknowledgments

Many thanks to those who have commented on or contributed to drafts including Elizabeth Goeddel, Steven Jones, Ion Juvina, James Kirk, Preeti Ramaraj, Bryan Stearns, and Robert Wray. Soar is the result of the contributions of dozens of researchers over the years. The references include many of those contributors but not all.

This work was supported by AFOSR under Grant Number FA9550-18-1-0168. The views and conclusions contained in this document are those of the authors and should not be interpreted as representing the official policies, either expressed or implied, of the Department of Defense or AFOSR. The U.S. Government is authorized to reproduce and distribute reprints for Government purposes notwithstanding any copyright notation hereon.

# Appendix A: Soar Systems

Our methodology is to test our theories in implementations and to inspire new theories and extensions by limitations we discover in those implementations. Below is a partial list of agents developed in Soar over the last 30 years, along with descriptions of the major research issues explored. The list is in reverse chronological order. Some of the text descriptions for pre-2005 systems are taken from Lehman et. al (2006).

**Atari (2022-): James Boggs**
Our goal for this project is to develop a single Soar agent that learns to play Atari games. We hypothesize that Soar can learn to play orders of magnitude faster than pure deep reinforcement approaches. This project is in its infancy, but the plan is to use a combination of general core knowledge about 2D games, the ability to extract game-specific knowledge from simple instructions, and the learning of internal action models of each game. Another key component is the extension of SVS to process the pixel-level game screens and the creation of a language that Soar agents can use to control the filters for extracting content from SVS.

**IBMEA (2020-): Steven Jones**
Instance-Based Means-Ends Analysis (IBMEA) is a general set of rules that enable an agent to learn action model knowledge needed for planning. As part of a broader theory for how to implement human-like event cognition, IBMEA uses episodic memory representations of previous events to learn action models. This knowledge is created as representations in working memory that are stored in semantic memory and that chunking uses to learn an associated operator in procedural memory. Using this learned knowledge (and prior action model knowledge), an agent then engages in means-ends analysis planning.

**MAST (2020-) Elizabeth Goeddel**
Mast (Motion planning with Agent Selection of Trajectory) is a motion planning approach for Soar agents that control robotic arms or other high-dimensional physical embodiments. Motion planning for such systems requires continuous, modality-specific processing and is most efficiently handled within SVS. The goal of MAST is to enable Soar agents to reason about actions in a way that is grounded to continuous motions in SVS but still integrated with symbolic knowledge in the rest of Soar. Following in the spirit of existing SVS spatial reasoning, MAST enables a Soar agent to interact with motor trajectories through the filter interface; symbolic information about movements is extracted into WM for agent reasoning. The agent can then compare a variety of trajectory options for an action and choose to execute the one that maximizes its task-level objectives, such as efficiency, caution, or visibility.

**Remote Associates Test (2018-): Jule Schatz**
The Remote Associates Test (RAT) is a task developed in 1962 by Mednick to measure convergent thinking skills in people (Mednick 1962). The task consists of a series of problems, where each problem consists of presenting the subject with a prompt word. The subject must determine a fourth solution word that is associated with all three prompt words. For example, if "swiss," "cake," and "cottage" are the prompt words, then "cheese" is the sought-after response. The Soar model of the RAT stores large preexisting knowledge bases of word associations in semantic memory, and use iterative retrievals combined with spreading activation to solve the RAT. The best model achieves a surprisingly good R2 of 0.98 and MSE of 0.29 when compared to human behavior in terms of problem difficulty (paper under review).

**Lucia (2017-): Peter Lindes**
Lucia (Lindes et al. 2017) is a model of language comprehension integrated into Rosie (see below). It is designed according to five qualitative principles including embodied, end-to-end comprehension; composable knowledge of the meaning of linguistic forms; incremental, immediate interpretation processing; using general cognitive mechanisms in Soar. It uses Embodied Construction Grammar as its method of representing composable knowledge of meaning and demonstrates that this knowledge can be processed incrementally using a novel comprehension algorithm that relies on the general cognitive mechanisms Soar to produce embodied, end-to-end comprehension.

**PROPS (2017-2021): Bryan Stearns**
PROPS is an agent that models human skill learning and transfer (Stearns et al., 2017), based on the PRIMs theory of skill composition (Taatgen, 2013). It can be given declarative descriptions of rules and then convert them into efficient procedural knowledge through practice in a manner that reproduces human learning curves. It has been applied to model humans in learning mental arithmetic tasks, computer text editor tasks, working memory training tasks, and various task-switching tasks (Stearns, 2021).

**Rosie (2011-)** is an Interactive Task Learning (ITL) agent that learns completely new tasks online from natural language instructions (Mohan et al., 2012; Mohan, 2015; Lindes et al., 2017), inspired by Instructo-Soar (Huffman, 1993). The emphasis with Rosie, and ITL, is to learn tasks quickly from situated interaction, generalizing from a single teaching interaction, without requiring many examples, in a similar manner to how humans learn new tasks from a teacher. Rosie has been embedded in simulation environments (April, AI2Thor), mobile robots (MagicBot, Fetch, Cosmo), and a desk-top robotic arm. When embodied in a robot, Rosie perceives its external environment through vision and range sensors and acts by issuing motor control commands. Rosie has learned to play over 60 different games and puzzles, such as Tic-tac-toe, river crossing puzzles, Solitaires, Sudokus, simple board games, and tile moving puzzles. For these games, Rosie learns all the elements of a task that enable it to understand and play the game: the goals, legal actions, failure conditions, and task-specific terminology, such as what 'captured' means in the context of the current game (Kirk, 2019). Rosie has also learned a large variety of tasks embodied in a mobile robot, such as navigation and delivery tasks. In its simulation environments for a mobile robot, it learns to perform patrolling tasks that involve visiting rooms, cleaning up, and raising an alarm if there is a (simulated) fire. These include both goal-based and procedural tasks, using diverse actions (physical, mental, and communicative) and diverse control structures (Mininger, 2021).

**Liar's Dice (2011)** is a Soar agent developed to play Liar's Dice, a competitive game involving hidden dice and round-robin bidding about the number of dice of a given face hidden under the cups of players (Laird et al. 2011). The agent employed simple models of its opponents as well as reinforcement learning to improve its play. A non-learning version was fielded as an iPhone app for many years, where Soar played against the user.

**MOUTBOT (2004)** is similar in spirit to TacAir-Soar but developed for military operations in urban terrain (Wray et al., 2005). In this system, Soar controlled individual adversaries who would defend a building against attack by military trainees. It required integration of reaction, planning, communication and coordination, and spatial reasoning.

**NTD-Soar (1994)** was a computational theory of the perceptual, cognitive, and motor actions performed by the NASA Test Director (NTD) as he utilizes the materials in his surroundings and communicates with others on the Space Shuttle launch team (Nelson, Lehman, & John, 1994). NTD-Soar combined NL-Soar and NOVA (a model of vision) with decision-making and problem-solving knowledge relevant to the testing and preparation of the Space Shuttle before it is launched.

**TacAir-Soar (1992-1997)** and **RWA-Soar (1995-1997)** were Soar models of human pilots used in training in large-scale distributed simulations (Jones et al., 1999; Tambe et al., 1995). They remain the largest Soar systems ever built (TacAir-Soar is currently > 8000 rules). They performed (in simulation) all of the U.S. tactical air missions for fixed-wing (jets and propeller planes) and rotary wing (helicopters) aircraft. The aircraft could fly the missions autonomously, dynamically changing the missions and the command structure, all the while communicating using standard military language. This work included early work on theories of coordination and teamwork that were realized in the STEAM system (Tambe, 1997).

**IMPROV (1995) Doug Pearson**
IMPROV is a computational theory of how to correct knowledge about what actions do in the world (Pearson & Laird, 1995). IMPROV combined SCA (a model of category learning) with knowledge about how to detect when the system's internal knowledge conflicts with what it observes in the world so that the system can automatically improve its knowledge about its interactions with the world.

**Instructo-Soar (1993) Scott Huffman**
Instructo-Soar embodied a computational theory of how people learn through interactive instruction (Huffman & Laird, 1995) and was a predecessor to Rosie. Instructo-Soar combined NL-Soar with knowledge about how to use and learn from instructions during problem solving to produce a system that learned new procedures through natural language.

# Appendix B: Glossary of Soar Terminology

**Chunking:** This is a learning mechanism that compiles the processing in a substate into a rule. That rule (a chunk) creates the same result of the substate, under the same conditions, but without the intermediate processing. Chunking is invoked whenever a result is created and creates a new rule that is immediately added to procedural memory.

**Decision procedure:** The decision procedure starts at the topstate and processes the preferences in the local state's preference memory until there is a change in the selected operator or impasse for a state. Within a state, it uses a fixed ordering of the preferences, starting with acceptable preferences. The remaining preferences filter out proposed operators until one of three conditions: there is a single operator, there are no operators remaining, or there are multiple proposed operators but no more preferences. The first of these results in the selection of that operator, whereas the latter two lead to impasses. When there are indifferent preferences, they are not considered until all symbolic preferences are considered, and at that point, they are used to make stochastic decisions biased by the numeric values, of the remaining operators.

**Episodic memory:** This long-term memory contains episodes, each of which is a snapshot of the topstate of working memory. An episode can be retrieved by providing a partial description of a state, and the most recent state that matches that cue will be retrieved. These episodes are temporally organized so that when one is retrieved, it is possible to retrieve either the next or previous episode.

**Impasse:** An impasse arises for a state when the decision procedure cannot select a new operator for the state. There are technically four types of impasses: state no-change, when no operator is proposed; operator no-change, when the same operator remains selected for two or more decisions; tie, when multiple operators are proposed and there are not sufficient preferences to select a single operator; conflict, when best/worst preferences conflict. When an impasse arises, a substate is created, with recursive problem solving. An impasse is resolved when a new decision can be made or when a different impasse arises. When an impasse is resolved all data structures linked solely to that state and not a superstate, are automatically removed from working memory.

**Operator:** An operator is created by an operator proposal rule that creates an acceptable preference and an associated data structure that describes the operator (as a graph structure). Only one operator can be selected for a state at a time. When an operator is selected, there is a special data structure added to working memory by the decision procedure, indicating that operator's selection. This structure is tested by operator application rules to ensure they apply only when the appropriate operator is selected.

**Preference:** A preference is a data structure created by a rule that is a statement as to the worthiness of an operator for selection for a specific state. When created, all preferences are added to preference memory. Acceptable preferences, which are statements that an operator can be considered for selection, are also added to working memory so that evaluation rules can match them and create other types of preferences. Symbolic preferences are either absolute (rejects, best, worst) or relative (better/worse, equal) statements of an operator or pair of operators. Numeric preferences contain the expected reward for an operator and are tuned by reinforcement learning.

**Preference Memory:** Contains all preferences that are currently active for a given state.

**Procedural memory:** This long-term memory contains all the agent's knowledge about skills, procedures, etc. All knowledge is represented as rules.

**Reinforcement Learning:** This is a learning mechanism that modifies the numeric indifferent preferences of evaluation rules based on reward. Soar uses supports many variants of reinforcement learning.

**Result:** A result is a working memory element (or set of connected elements) that are created by rules testing structures in a substate but is linked to a superstate. It is the basis for creating a chunk.

**Rule:** Each rule contains a list of conditions and a list of actions. The conditions are matched against working memory and the actions either create preferences (in preference memory) or change working memory. Rules have four different functions: elaboration, operator proposal, operator evaluation, and operator application. A given rule should not mix functions. During the elaboration phase of the decision cycle, all matching elaboration, operator proposal, and operator evaluation rules fire in parallel. These rules retract their actions, which create either preferences or working memory elements, when the conditions no longer match. During the application phase, all operator application rules fire in parallel to change working memory – they do not create preferences. Application rules are defined as rules that test the current operator and state and modify the state. These rules do not retract their actions.

**Semantic memory:** This long-term memory contains concepts, including facts and similar information as symbolic graph structures. A concept is retrieved through the creation of a cue in the semantic memory cue buffer of working memory. Semantic memory uses a combination of base-level activation and spreading activation to determine which concept to retrieve.

**State (and substate):** The structures in working memory are organized into states. The topstate always exists and a substate is created when there is an impasse in decision making. There can be only one substate per state and substates can have their own substates from an impasse in operator selection. There is a link from each substate to its superstate, but not in the other direction. Thus, the most recent state/substate is the root of all structures in working memory.

**Topstate:** The topstate always exists in working memory and contains buffers for semantic memory, episodic memory, SVS, other perceptional systems, and the motor system. It is the only state through which external perception is available and the only state in which motor actions can be created. It is also the only state that is captured by episodic memory.

**Working memory:** This short-term memory contains the agent's representation of its current situation. Working memory is represented as a graph structure, organized as a topstate and substates. The most recent substate is the root, with a link to the next most recent substate, which repeats to the top-level state. The topstate has buffers for interacting with episodic memory, semantic memory, the motor system, SVS, and any other perceptual module. Substates have buffers for episodic memory and semantic memory, as well as an purely local version of SVS.